\documentclass[accepted]{uai2026} 
                        
\usepackage{microtype}
\usepackage{enumitem}
\usepackage{graphicx}
\usepackage{amsmath,amsfonts,bbm}
\usepackage{subcaption}
\usepackage{booktabs} 
\usepackage{tikz}
\usetikzlibrary{calc,positioning, arrows.meta, backgrounds, fit, shadows,decorations.pathreplacing}

\usepackage[table]{xcolor}
\usepackage{hyperref}

\usepackage{amsmath}
\usepackage{amssymb}
\usepackage{mathtools}
\usepackage{amsthm}
\usepackage{nicematrix, tikz}
\usepackage[capitalize,noabbrev]{cleveref}

\theoremstyle{plain}
\newtheorem{theorem}{Theorem}[section]

\theoremstyle{definition}
\newtheorem{definition}[theorem]{Definition}

\theoremstyle{remark}

\usepackage[american]{babel}

\usepackage{fontawesome5}
\usepackage{natbib} 
\usepackage{mathtools} 
\usepackage{booktabs} 
\usepackage{tikz} 

\usepackage[table]{xcolor}
\usepackage{pgfmath}
\usepackage{etoolbox}
\usepackage{booktabs}

\colorlet{high}{purple!70!black}
\colorlet{low}{yellow!60}

\colorlet{high}{teal!80!black}
\colorlet{low}{brown!20!white}

\colorlet{high}{blue!80!black}
\colorlet{low}{purple!30!white}

\def\opacity{30}

\newcommand{\applygradient}[5]{%
    \pgfmathparse{int(round(100*(#1-#4)/(#5-#4)))}%
    \xdef\tempcolor{\pgfmathresult}%
    \cellcolor{high!\tempcolor!low!\opacity}%
    \ifnum\tempcolor>70\color{white}\else\color{black}\fi%
    \ifstrequal{#3}{bold}{\textbf{#1}}{#1}%
    \if\relax\detokenize{#2}\relax\else\;\scriptsize #2\fi
}

\title{Recursive Inference Machines for Neural Reasoning} 
\author[1]{Mieszko Komisarczyk$^*$}
\author[1]{Saurabh Mathur$^*$}
\author[1]{Maurice Kraus}
\author[2]{Sriraam Natarajan}
\author[1,3,4]{Kristian Kersting}
\affil[1]{%
    Department of Computer Science, Technical University of Darmstadt, Germany
}
\affil[2]{%
    Department of Computer Science, The University of Texas at Dallas, USA
}
\affil[3]{
    Hessian Center for Artificial Intelligence (hessian.ai), Darmstadt, Germany 
}

\affil[4]{
    German Research Center for AI (DFKI)
}
\begin{document}
\maketitle
\def\thefootnote{*}\footnotetext{Equal contributors.}\def\thefootnote{\arabic{footnote}}
\begin{abstract}
Neural reasoners such as Tiny Recursive Models (TRMs) solve complex problems by combining neural backbones with specialized inference schemes. Such inference schemes have been a central component of stochastic reasoning systems, where inference rules are applied to a stochastic model to derive answers to complex queries. 
In this work, we bridge these two paradigms by introducing Recursive Inference Machines (RIMs), a neural reasoning framework that explicitly incorporates recursive inference mechanisms inspired by classical inference engines. We show that TRMs can be expressed as an instance of RIMs, allowing us to extend them through a reweighting component, yielding better performance on challenging reasoning benchmarks, including ARC-AGI-1, ARC-AGI-2, and Sudoku Extreme. Furthermore, we show that RIMs can be used to improve reasoning on other tasks, such as the classification of tabular data, outperforming TabPFNs.
\end{abstract}

\section{Introduction}\label{sec:intro}
Neural reasoners have garnered considerable interest owing to their impressive performance on challenging reasoning benchmarks. These models solve reasoning tasks by exploiting patterns in their training data, allowing them to generate answers to problems for which exact symbolic reasoning is intractable~\citep{shi2023satformer}. However, this also limits them to problems of complexity similar to that of their training data; they struggle to generalize to problems requiring longer horizons. Moreover, learning models to solve more complex problems requires deeper architectures. 

To mitigate these limitations, recent work has explored test-time scaling approaches that dynamically allocate additional computation during inference. Techniques such as chain-of-thought prompting~\citep{wei2022chain}, self-verification~\citep{wang2022self}, and latent reasoning~\citep{wang2025hierarchical,jolicoeur2025less} allow a fixed neural backbone to perform multi-step reasoning, leading to improved performance on long-horizon tasks. Despite their empirical success, these methods are largely introduced as heuristic procedures, lacking a unifying formal framework that explains why they work or how they can be systematically composed and extended.

However, answering complex queries by reasoning about correlations between patterns has long been studied in probabilistic inference. Approximate inference methods such as belief propagation \citep{pearl2022reverend} and Gibbs sampling \citep{geman1984stochastic} decompose a long-horizon reasoning task into a sequence of smaller computational blocks; each block is learned or adapted separately.

We argue that modern test-time neural reasoning procedures can be understood through a similar lens. Specifically, we hypothesize that many such procedures can be naturally expressed as programs in a stochastic programming language, where neural components implement local inference operators and control flow specifies their recursive interaction. This perspective provides a unified semantics for existing methods, while enabling modularity, compositionality, and reuse across reasoning systems.

To this end, we introduce Recursive Inference Machines (RIMs), a framework for unified neural reasoning. RIMs make test-time reasoning explicit, allowing complex reasoning behavior to emerge from repeated applications of simple, reusable modules.

Specifically, we make the following key contributions: 
\begin{itemize}
\item We present RIMs as a general framework for neural reasoning architectures. 
\item 
We empirically demonstrate that RIMs outperform Tiny Recursive Models (TRMs, \citet{jolicoeur2025less}) on challenging benchmarks such as Sudoku Extreme and ARC, as well as TabPFN~\citep{hollmann2025accurate} on medical diagnosis benchmarks under heavy observational noise.
\end{itemize}

\section{Background}
Recursive Inference Machines are related to different lines of research.

\textbf{Neural Reasoning.}

Neural reasoning addresses reasoning tasks by training or adapting neural networks to perform multi-step inference directly from data. Unlike classical symbolic reasoners that rely on rule-based search, neural reasoners exploit learned latent representations to approximate reasoning behavior~\citep{bhuyan2024neuro}. These approaches have achieved strong empirical performance on large-scale benchmarks where symbolic pipelines are often brittle, as seen in models like SATFormer, which learns effective heuristics for satisfiability problems~\citep{shi2023satformer}.

Despite these successes, standard feedforward and Transformer architectures struggle with long-horizon reasoning due to their fixed computational depth. This constraint effectively limits the number of sequential steps in a single forward pass, preventing them from accurately solving tasks that require intricate algorithmic logic.

To overcome these depth constraints, recent work has focused on test-time scaling, allocating additional computation during inference via specialized schemes. These include chain-of-thought prompting \citep{wei2022chain} and iterative self-verification or refinement \citep{liu2025trust}. While such mechanisms allow for arbitrarily complex computations in theory \citep{schuurmans2024autoregressive}, they often lack a systematic interpretation of their underlying "thinking" strategies, making the process difficult to improve.

\textbf{Hierarchical Reasoning Model and Tiny Recursive Model.} A more structured approach to iterative computation is found in recurrent architectures like Hierarchical Reasoning Models (HRMs,\citet{wang2025hierarchical}) and Tiny Recursive Models (TRMs, \citet{jolicoeur2025less}) that update latent states over multiple cycles.

HRMs employ two interdependent recurrent modules operating at different temporal scales. A low-level module executes rapid, local refinement until reaching a local equilibrium, at which point a high-level module provides a new abstract context to restart the process. While HRMs achieve competitive performance on benchmarks like ARC-AGI and Sudoku Extreme, parts of the method are motivated by arguments invoking the Implicit Function Theorem \citep{krantz2002implicit}, whose underlying assumptions have been questioned in practical settings \citep{jolicoeur2025less}.

TRMs simplify the HRM architecture into a single recurrent unit using two nested loops. The latent states are interpreted more naturally: a short-term "scratchpad" $z_L$ captures intermediate states, while $z_H$ represents the evolving global solution~\citep{asadulaev2025deep}. By unrolling recursive steps and employing deep supervision, TRMs outperform significantly larger models with only 7M parameters~\citep{roye2025tiny}.

Whether through dual-speed modules (HRM) or nested loops (TRM), these models implement a form of sequential refinement. However, they lack a systematic interpretation of their latent state updates, making their reasoning strategies difficult to interpret or improve.

\textbf{Probabilistic Inference.} Reasoning with uncertain models has long been studied in the field of probabilistic inference. Probabilistic inference algorithms aim to answer queries, such as marginal statistics over unobserved variables given observed data. While exact inference is tractable for specialized structures like trees, it remains intractable for general probabilistic models, motivating several approximate inference methods~\citep{pearl2014probabilistic, koller2009probabilistic}.

Approximate inference methods scale reasoning to high-dimensional spaces by employing iterative local computations to approach the true posterior distribution. In this work, we focus on two such methods: Sequential Monte Carlo (SMC) and Gibbs Sampling.

SMC represents the posterior as a set of weighted samples (particles) that evolve through a sequence of proposal and correction steps. A proposal distribution generates new candidate states, which are then adjusted by importance weights. This weighting mechanism is essential for correcting the bias introduced by the proposal. In Section~\ref{section_3}, we show that recursive models like TRMs effectively implement the proposal but omit the weighting term, leading to sub-optimal reasoning trajectories.

Gibbs sampling approximates complex posteriors by iteratively updating each variable conditioned on all others variables. This iterative process allows the sampler to explore the joint distribution by making local, conditional moves. We use this perspective to adapt tabular transformers to datasets with systematic noise.

Using these stochastic inference strategies, we aim to provide a principled foundation for neural reasoning by interpreting the neural reasoners' thinking steps as samples approximating a posterior. However, rather than using hand-crafted distributions, we follow the Inference Machine paradigm to learn these transitions directly from data.

\textbf{Inference Machines.} While stochastic inference provides a rigorous framework for reasoning under uncertainty, learning explicit generative models remains challenging in complex domains. The Inference Machine (IM) paradigm addresses this by forgoing the strict separation between models and inference methods~\citep{ross2011learning}. Instead of learning a general-purpose model, this paradigm treats the neural architecture as a parameterization of 
reasoning, recasting it as a sequence of optimized predictions.

By merging the model and the inference algorithm into a single machine, we can learn to answer specific queries directly. For example, Message-Passing Inference Machines (MPIMs, \citep{ross2011learning}) replace the disconnected components of a graphical model and the belief propagation algorithm with a set of regression modules. These modules learn to perform inference by predicting messages based on local neighborhood context, thereby amortizing the cost of reasoning. Consequently, at test time, the model replaces expensive global optimizations with a fixed sequence of local operations compiled to navigate the specific manifolds of the target task.

We generalize this paradigm toward what we call Recursive Inference Machines (RIMs). We argue that architectures like TRMs are effectively approximate Inference Machines for SMC, parameterized with highly expressive neural backbones. Generalizing TRMs this way allows one to identify missing components, such as the importance Reweighter, and refine the architecture accordingly. Moreover, as we will demonstrate, the Inference Machine paradigm extends naturally to other algorithms and in-context learning. This enables one to adapt, for instance, a pretrained tabular transformer (TabPFN) within a RIM to realize a Gibbs sampler that learns to adapt to noisy domains.

\section{Recursive Inference Machine}
\label{section_3}


We now introduce Recursive Inference Machines (RIMs), a unified framework for neural reasoning. RIMs cast neural reasoners' latent state updates as a sequence of transitions defining a learned inference machine. 

\begin{definition}
    We define a Recursive Inference Machine as the tuple $\langle x, y^{(0)}, z^{(0)}, G, S, \mathcal{R}\rangle,$ where 
\begin{itemize}
    \item $x$ is the problem description;
    \item $y^{(0)}, z^{(0)}$ are the initial solution and the inital state, respectively;
    \item $S$ is the Solver, which proposes an update to the state conditioned on the current solution, the previous state and the problem description;
    \item $G$ is the Generator, which generates a candidate update to the solution, conditioned on all the updates to the state and solution;
    \item $\mathcal{R}$ is the Reweighter, which performs the actual updates to the state and the solution by weighing their current values against candidate updates.
\end{itemize}
\end{definition}

\begin{figure}[t]
    \centering

    \includegraphics[width=.9\linewidth]{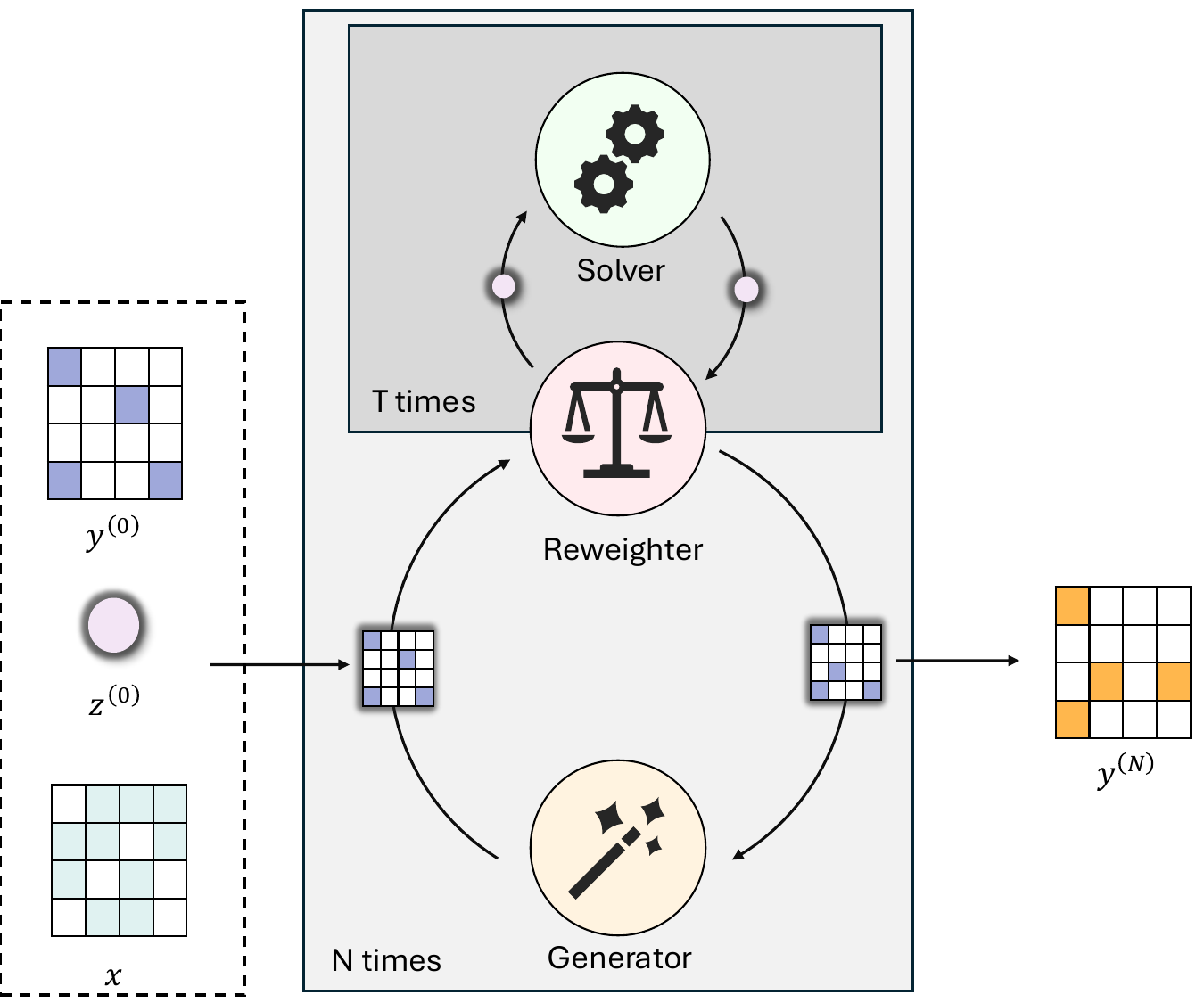} 
    
    \caption{{\bf A Recursive Inference Machine (RIM)} that solves problems described by $x$, given an initial solution $y^{(0)}$, and an initial state ($z^{(0)}$). It consists of a \colorbox{green!15}{\textbf{Solver}}, a \colorbox{red!15}{\textbf{Reweighter}}, and a \colorbox{orange!15}{\textbf{Generator}}. The components solve the problem by alternating between the Solver updating the state recursively for $T$ steps, and the Generator using these state-updates to generate an updated solution. After repeating this $N$ times, the RIM produces the solution $y^{(N)}.$}
    \label{fig:rim_gen} 
\end{figure}

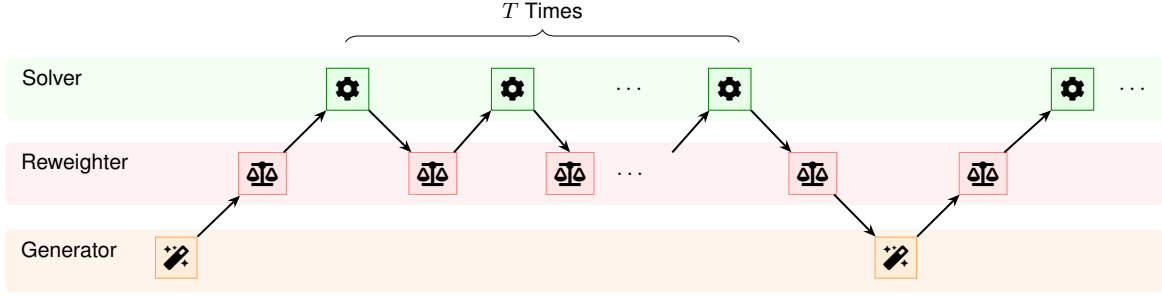
\begin{figure*}[ht]
    \centering
\resizebox{.9\textwidth}{!}{
\begin{tikzpicture}[
  x=0.015cm, y=-0.015cm,
  arr/.style={-{Stealth[length=5pt,width=4pt]}, thick},
]

\def\s{40} 

\fill[green!5,  rounded corners=3pt] (1,94)  rectangle (1090,154);
\fill[red!5,    rounded corners=3pt] (0,174) rectangle (1090,234);
\fill[orange!7.5, rounded corners=3pt] (0,257) rectangle (1090,317);

\node[anchor=north west,font=\small\sffamily] at (8,  97)  {Solver};
\node[anchor=north west,font=\small\sffamily] at (7,  177) {Reweighter};
\node[anchor=north west,font=\small\sffamily] at (7,  260) {Generator};

\tikzset{
  Sbox/.style={draw=green!50!black, fill=green!10,  minimum size=17pt},
  Rbox/.style={draw=red!50,         fill=red!10,    minimum size=17pt},
  Gbox/.style={draw=orange!70,      fill=orange!15, minimum size=16pt},
}
\node[Sbox] (S1) at (282+\s, 124) {\faCog};
\node[Sbox] (S2) at (437+\s, 124) {\faCog};
\node[Sbox] (S3) at (641+\s, 124) {\faCog};
\node[Sbox] (S4) at (963+\s, 124) {\faCog};

\node[Rbox] (R1) at (202+\s, 204) {\faBalanceScale};
\node[Rbox] (R2) at (362+\s, 204) {\faBalanceScale};
\node[Rbox] (R3) at (491+\s, 204) {\faBalanceScale};
\node[Rbox] (R4) at (719+\s, 204) {\faBalanceScale};
\node[Rbox] (R5) at (879+\s, 204) {\faBalanceScale};

\node[Gbox] (G1) at (121+\s, 284) {\faMagic};
\node[Gbox] (G2) at (797+\s, 284) {\faMagic};

\node[font=\small] at (548+\s, 124) {$\dots$};
\node[font=\small] at (549+\s, 204) {$\dots$};
\node[font=\small] at (1021+\s,124) {$\dots$};

\draw[decorate, decoration={brace, amplitude=5pt, raise=2pt}]
   (280+\s, 84) -- (650+\s, 84) 
  node[midway, above=8pt, font=\small] {$T\textsf{ Times}$};

\draw[arr] (141+\s, 264) -- (181+\s, 225);  
\draw[arr] (222+\s, 184) -- (262+\s, 144);  
\draw[arr] (302+\s, 144) -- (342+\s, 184);  
\draw[arr] (382+\s, 184) -- (417+\s, 144);  
\draw[arr] (457+\s, 144) -- (491+\s, 184);  
\draw[arr] (587+\s, 184) -- (622+\s, 144);  
\draw[arr] (661+\s, 144) -- (699+\s, 184);  
\draw[arr] (739+\s, 224) -- (778+\s, 264);  
\draw[arr] (817+\s, 264) -- (857+\s, 224);  
\draw[arr] (899+\s, 184) -- (943+\s, 144);  

\end{tikzpicture}

}

    \caption{\textbf{Unrolling a Recursive Inference Machine (RIM).} The inner loop (indexed by $j=1,\dots,T$) acts as the \colorbox{green!15}{\textbf{Solver}}, which extends the short-term reasoning "scratchpad" $z^j$ by proposing new states $\hat{z}^{j+1}$ based on the problem description $x$ and current solution $y^i$. The outer loop (indexed by $i=1,\dots,N$) implements the  \colorbox{orange!15}{\textbf{Generator}}, which updates the solution $y^i$ after each phase of local refinement. The \colorbox{red!15}{\textbf{Reweighter}} intercepts the updates proposed by the Solver and Generator and weighs them against their previous values, stabilizing the trajectory. 
    }
    \label{fig:rim_unrolled}
\end{figure*}

Fig.~\ref{fig:rim_gen} visualizes the general structure of a RIM. A RIM solves the problem described by $x$ by alternating between recursive state updates and solution updates. The Solver, along with the Reweighter, recursively generates a sequence of $T$ state updates. Each state update is a combination of the Solver proposing an update $\tilde z^{(i)}$ and the Reweighter weighing it against $z^{(0)},\dots,z^{(i-1)}$ to produce $z^{(i)}$. This results in state updates $z^{(1)},\dots, z^{(T)}.$ The Generator uses this sequence to produce a candidate solution $\tilde y^{(1)},$ which the Reweighter processes to to obtain the updated solution $y^{(1)}.$ This entire process is repeated $N$ times, yielding  the final solution $y^{(N)}$. 

Fig.~\ref{fig:rim_unrolled} visualizes the operation of a RIM by unrolling it. It starts with the initial solution $y^{(0)}$ and initial state $z^{(0)}$. These are used to propose a candidate updated state $\tilde z^{(1)}$, which is fed to the Reweighter to obtain the updated state $z^{(1)}$; this is repeated to generate a sequence of T states $z^{(1)},\dots,z^{(T)}.$ The Generator uses this sequence of states to generate a candidate updated solution $\tilde y^{(1)},$ which is fed to the Reweighter to obtain the updated solution $y^{(1)}.$ This process of generating a sequence of state updates and solution updates is repeated $N$ times, yielding a sequence of updated solutions $y^{(1)}, \dots, y^{(N)};$ the last solution is returned as the output. 

This unrolling process is a formal generalization of Sequential Monte Carlo (SMC) in reasoning space. A RIM generates a trajectory by starting with $y^{(0)}, z^{(0)},$  extending the state through $T$ steps of the Solver, and then resampling a new solution $y^{(1)}$ via the Generator and Reweighter. 

This mapping allows us to interpret the neural latent states not merely as feature vectors, but as sufficient statistics parameterizing unnormalized belief distributions over the reasoning state. In this framework, the recursive update functions (the Solver and Generator) act as learned transition operators trained to move the latent belief toward regions of high posterior probability.

In the subsequent sections, we consider various instantiations of RIMs.

\section{Family of Recursive Inference Machines}
The RIM framework serves as a generative template for reasoning architectures, where specific functional choices for the Solver, Generator, and Reweighter define the resulting model's inductive biases. In this section, we explore this family of machines by formalizing existing neural models within the RIM framework and introducing new instantiations designed for effective reasoning.
\begin{table*}[ht!]
\centering
\resizebox{.7\linewidth}{!}{\begin{tabular}{@{}l >{\columncolor{green!7.5}}l >{\columncolor{orange!7.5}}l >{\columncolor{red!7.5}}l ll@{}}
\toprule
\textbf{Name} &  \textbf{Solver (\faCog)} & \textbf{Generator (\faMagic)} & \textbf{Reweighter (\faBalanceScale)}  \\ \midrule
SimRIM (D) &  $f_L(\Theta_L)$ & $f_H(\Theta_H)$ & $\mathbbm{1}$ (Identity) \\
SimRIM (S) & $f(\Theta)$ & $f(\Theta)$ & $\mathbbm{1}$ (Identity)  \\
RIMA & $f(\Theta)$ & $f(\Theta)$ & EMA \\
RIMFormer & $f(\Theta)$ & $f(\Theta)$ & Transformer  \\
{TabRIM} & $P_\text{TabPFN}(X_i \mid x_{-i}, y,  D_\text{train})$ & $P_\text{TabPFN}(Y \mid x, D_\text{train})$  & $P(e \mid \hat{x})$ \\ \bottomrule
\end{tabular}} 
\caption{\textbf{Family of RIMs.} We categorize the architectures based on their components: \colorbox{green!15}{\textbf{Solver}}, \colorbox{orange!15}{\textbf{Generator}}, and \colorbox{red!15}{\textbf{Reweighter}}.  The two variants of Simple RIM (SimRIM, aka TRM), Decoupled (D) and Shared (S), have a simple identity function Reweighter. They are identical to HRM and TRM, respectively. RIMA and RIMFormer improve upon SimRIM (S) by employing more expressive Reweighters. TabRIM implements Solver and Generator via TabPFN forward passes, while the Reweighter encodes prior knowledge about noise.
}
\end{table*}

\subsection{Learning to reason}
We conceptualize reasoning as the iterative refinement of a latent state through a structured Solver-Generator-Reweighter loop. To demonstrate the flexibility of our framework, we now describe how specific choices of these components recover and extend existing neural reasoning architectures.

\textbf{SimRIM.} We first instantiate our framework for recursive reasoning architectures by introducing the \textit{Simple RIM (SIMRIM)}. This instantiation provides a unifying formalism that encompasses models such as Hierarchical Reasoning Models (HRM)~\citep{wang2025hierarchical} and Tiny Recursive Models (TRM)~\citep{roye2025tiny}. 

We define SimRIM as the RIM 

$\langle x, z_H^{(0)}, z_L^{(0)}, G, S, \mathcal{R} \rangle,$ where $x$ is the problem description embedding, $z_H^{(0)}$ and $ z_L^{(0)}$ are the initial solution and initial state, and
\begin{itemize}

      \item The Solver $S$ is implemented by $f_L,$ performing $T$ state updates before every reasoning state update $z_L^i = f_L(z_L^{i-1}, z_H^j, x)$;

    \item The Generator $G$ is implemented by $f_H,$ performing $N$ solution updates $z_H^j = f_H(z_H^{j-1}, z_L^{T-1})$;

    \item The Reweighter $\mathcal{R}$ is the identity function.
\end{itemize}
Note that SimRIM is identical to the HRM. 
Moreover, when the Solver and Generator networks, $f_L$ and $f_H,$ are implemented using the same neural backbone (say, $f$), the above reduces to a TRM. In both cases, the Reweighter is the identity function. So, without a non-trivial identity function, the model starts with the initial solution and state $\langle z_H^{(0)},z_L^{(0)}\rangle$ and extends it by appending $z_L^{(i)}$ for T steps and directly generates a solution update $z_H^{(1)};$ this process is repeated $N$ times.

\textbf{RIMA.} The observation that existing models employ an identity reweighting function motivates a more expressive parameterization. In SMC, the Reweighter is critical for correcting proposal bias and preventing reasoning drift.

On the other hand, moving averages are ubiquitous and can be found in GRU~\citep{cho2014gru}, LRU~\citep{ahuja2018lru}, MAMBA~\citep{gu2024mamba}, and RWKV~\citep{peng2023rwkv}. 
For RIMs, exponential moving averages offer an elegant way to balance the past and the present. Rather than treating all history equally or discarding it abruptly, as the SimRIMs currently do, we can naturally down-weight older information while keeping it in view using exponential moving averages. 

That is, the more distant an intermediate result ("thought"), the less it influences the current "thought". This makes RIM use a principled and computationally lightweight mechanism for tracking changing thoughts: recent observations and thoughts shape the current thought most strongly, yet the full history leaves a trace.

The decay parameters give intuitive control over this trade-off, from long-horizon memory to rapid adaptation, making EMA a versatile building block of RIM $\langle x, z_H^{(0)}, z_L^{(0)}, G, S, \mathcal{R} \rangle$. This leads us to the first proposed model - RIMA, which has the following dynamics. 
\begin{itemize}
    \item Both the Solver $S$ and Generator $G$ are defined similar to the SimRIM with $f_H=f_L$, producing updates for $\tilde z_L^{(i)}$ and $\tilde z_H^{(j)}$;
    \item The Reweighter $\mathcal{R} = \left(\mathcal{R}_L, \mathcal{R}_H \right)$ contains two neural backbones. The $\mathcal{R}_L$ is defined as
\begin{eqnarray*}
\mathcal{R}_L \left(\tilde z_L^{(i)}, z_L^{(i)} \right) &= \alpha^{(i)}_L \tilde z^{(i)}_L + \left(1-\alpha^{(i)}_L \right)z^{(i)}_{L},
\end{eqnarray*}
where $\alpha^{(i)}_L$ controls the update inertia. This coefficient can be a fixed scalar, learnable scalar, or vector, computed via input-dependent gating:
\begin{eqnarray*}
\alpha^{(i)}_L &= \sigma \left( \text{LinLayer}\left(\tilde z^{(i)}_L\right)  \right).
\end{eqnarray*}
The $\mathcal{R}_H$ is defined analogously.
\end{itemize}

From now on, we call a Reweighter \textit{dynamic} if it contains learnable parameters, and \textit{static} otherwise.

\textbf{RIMformer.} For reasoning tasks involving long horizons and frequent backtracking, maintaining a deeper historical context is essential.
To this end, we introduce the $k$-lookback Reweighter $\mathcal{R},$ a function that conditions the update on the current candidate and $k$ previous values:
$$\mathcal{R}: 
\underbrace{\mathbb{R}^d \times \mathbb{R}^d \times \cdots \times \mathbb{R}^d}_{k+1 \text{ times}} \to \mathbb{R}^d$$

where $d$ denotes the dimension of the hidden space. 

Formally, given a candidate output $\tilde z^{(n)}_X$ from the backbone, the update is defined as

\begin{eqnarray*}
    z^{(n+1)}_X = \mathcal{R}_X \left( \tilde  z^{(n)}_X, z^{(n)}_{X}, \ldots, z^{(n-k)}_{X} \right), 
\end{eqnarray*}
where $X \in \{L, H \}$ and $n \in \{1, \dots, Y_X \}$, where $Y_L=T$ and $Y_H=N$.  

The lookback mechanism serves as the primary motivation for our proposed architecture, the RIMformer. In this model, the Reweighter $\mathcal{R}$ is implemented as a Transformer block, which explicitly captures the dependencies across the entire reasoning history via its self-attention mechanism. Formally, we define the RIMformer as the tuple $\langle x, z_H^{(0)}, z_L^{(0)}, G, S, \mathcal{R} \rangle$, where
\begin{itemize}
    \item The Solver $S$ and Generator $G$ are defined as for the RIMA;
    \item The Reweighter $\mathcal{R} = \left(\mathcal{T}_L, \mathcal{T}_H \right)$ consists of two transformer models. The $\mathcal{T}_L$ is defined as follows
    \begin{equation*}
        \mathcal{T}_L \left(h_L \right) = \text{MLP } \left( \text{Norm}\left( A' \right) \right) + A',
    \end{equation*}
    where $h_L=\left( \tilde z^{(i)}_L,  z^{(i)}_{L}, \ldots, z^{(0)}_{L} \right)$ and $A'$ denotes the attention output representations corresponding to the candidate tokens after applying self-attention over the concatenated history and candidate sequence. 
    The definition of $\mathcal{T}_H$ is analogous. 
\end{itemize}

\begin{figure}[ht!]
    \centering 
\resizebox{.85\linewidth}{!}{
\begin{tikzpicture}[
    node distance=2cm and 1cm,
    observed/.style={circle, draw=teal!80!black, fill=teal!10, thick, minimum size=1.2cm, font=\large, drop shadow={opacity=0.1}},
    target/.style={circle, draw=orange!80!black, fill=orange!30, ultra thick, minimum size=1.2cm, font=\large, drop shadow={opacity=0.1}},
    latent/.style={circle, draw=violet!80!black, fill=violet!5, thick, minimum size=1.2cm, font=\large},
    edge/.style={->, >={Stealth[length=3mm]}, thick, draw=black!70},
    dep/.style={<->, thick, dashed, draw=cyan!50!black!60} 
]

    \node[latent] (X1) {$X_1$};
    \node (T) [right=of X1] {$\dots$};
    \node[latent] (X2) [right=of T] {$X_n$};
    \node[target] (Y) [right=of X2] {$Y$};
    
    \node[observed] (E1) [below=of X1] {$E_1$};
    \node[observed] (E2) [below=of X2] {$E_n$};

    \draw[edge] (X1) -- (E1);
    \draw[edge] (X2) -- (E2);

    \draw[edge] (X1) to [bend left=30] (X2);
    \draw[edge] (X2) to [bend left=30] (Y);
    \draw[edge] (X1) to [bend left=45] (Y);

    \draw[edge] (X2) to [bend left=30] (X1);
    \draw[edge] (Y) to [bend left=30] (X2);
    \draw[edge] (Y) to [bend left=45] (X1);

    \begin{scope}[on background layer]

        \node[fill=violet!2, draw=violet!15, rounded corners=10pt, 
              fit=(X1) (Y), 
              inner ysep=30pt, inner xsep=10pt,
              label={[anchor=south east, font=\itshape\color{violet!50!black}, yshift=2pt]}] (plateL) {};
        
        \node[fill=teal!5, draw=teal!20, rounded corners=10pt, 
              fit=(E1) (E2), 
              inner ysep=15pt, inner xsep=10pt,
              label={[anchor=north east, font=\itshape\color{gray}, yshift=-2pt]south east:}] (plateE) {};
    \end{scope}

    \node[left=0.8cm of X1, font=\sf\bfseries\color{violet!70!black}] {Latent};
    \node[left=0.8cm of E1, font=\sf\bfseries\color{teal!70!black}] {Observed};

    \node[observed] (E1) [below=of X1] {$E_1$};
    \node[observed] (E2) [below=of X2] {$E_n$};

\end{tikzpicture}

}

    \caption{\textbf{Tabular RIM (TabRIM) for Reasoning under Uncertainty}
    To reason about the distribution over the target $Y$ given noisy observations $E=e,$ the TabRIM decomposes the task into a Solver-Reweighter-Generator loop. The {\bf Solver} iteratively denoises the observation by iteratively resampling latent variables from their full conditionals. Each such conditional is implemented as a single forward pass over TabPFN. The {\bf Reweighter} weighs these samples against the observed evidence to ensure consistency with it. These consistency-weighted samples are used by the {\bf Generator} to infer the final target distribution by computing the empirical expectation over them.}
   
    \label{fig:tabpfn-rim}
\end{figure}
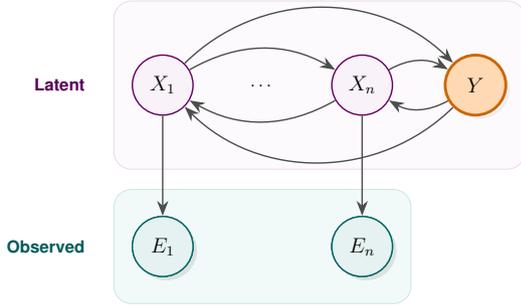
\subsection{In-context reasoning with TabPFN}
To highlight the applicability of the RIM framework to pretrained models, we consider tabular reasoning with TabPFN due to \cite{hollmann2025accurate}. TabPFN is a prior-fitted transformer, which has been trained on a large number of synthetic datasets; it answers predictive queries without an explicit learning step by treating the training data as in-context examples. While this allows TabPFN to exploit prior knowledge embedded in its weights, the lack of an explicit model makes it challenging to deal with noise at deployment. Deployed models must often process data that is significantly noisier than their training data, which is typically collected from controlled environments.

In traditional machine learning, models are made robust to deployment-time noise by integrating domain-specific knowledge during the training phase, such as by augmenting the loss function with a knowledge-based regularization term~\citep{karpatne2022knowledge,karanam2025unified}. However, exploiting such domain knowledge is uniquely challenging for TabPFNs because they lack an explicit gradient-based learning step. To enable TabPFNs to reason about noisy input features $E$ to infer the distribution over the target $Y,$ we construct the following RIM and call it Tabular RIM (TabRIM):

\textbf{Problem description} consists of a training context $D_\text{Train} = \{(x^{(i)}, y^{(i)})\}_{i=1}^N$ and a set of noisy test features $e$

\textbf{Latent state} consists of a sequence of samples ${\hat{x}}$ drawn from a Markov Chain approximating the distribution over denoised features, $P(X \mid e, D_\text{Train})$. Each sample represents a potential denoised version of the input. 

\textbf{The Solver} is responsible for state refinement. It uses TabPFN's in-context learning to generate denoised samples of the latent features. By iteratively sampling from the full conditionals over each feature, the Solver produces a set of clean feature hypotheses $\hat{x}$ consistent with the training data's distribution.

\textbf{The Generator} infers the final answer by producing the predictive distribution over the target $Y$. It marginalizes over the denoised samples provided by the Solver:
$$P(Y \mid e) = \frac{1}{Z|H|} \sum\nolimits_{\hat{x} \in H} w(\hat{x}, e)P(Y \mid X=\hat{x}, D_\text{Train}),$$
where $Z = \sum\nolimits_{\hat{x} \in H} w(\hat{x}, e). $

\textbf{The Reweighter} assigns importance weights $w(\hat{x}, e)$ to the Solver’s denoised samples to account for their likelihood given the observed evidence. Given a feature-wise deviation rate $\epsilon$, the weight of a sample can be defined using a function $w: \text{Domain}(X)^2 \mapsto \mathbb{R}^+$. For example, a simple Reweighter could be defined as:
$$w(\hat{x}, e) = \prod\nolimits_{i=1}^{n} \epsilon \cdot\mathbbm{1}[e_i \neq \hat{x}_i] + (1-\epsilon) \cdot\mathbbm{1}[e_i = \hat{x}_i]$$
where $\epsilon > 0$ is a known feature-wise noise parameter.

This reweighting ensures that the Generator's predictive distribution is dominated by the predictions corresponding to the most plausible denoised features, allowing it filter out the effect of noise.

The efficacy of the Tabular RIM stems from its formal equivalence to Gibbs sampling from a probabilistic generative model over the observed features $E,$ their clean version $X,$ and the target $Y.$
$$P(E, X, Y) \approx {P(E \mid X)} {P(Y \mid X)} {\prod\nolimits_i P(X_i \mid X_{-i}, Y)} $$
The structure of this model is presented in Fig.~\ref{fig:tabpfn-rim}
The operation of TabRIM is identical to inferring the following expectation using Gibbs sampling.
$$P(Y \mid E=e) = \mathbb{E}_{x \sim P(X\mid e)}[P(Y \mid X=x)]$$
The Solver iteratively samples from TabPFN-based full conditionals to generate samples $\{\hat{x}^{(k)}\} \sim P(X),$ which are reweighed using the emission probability of evidence $e$ as $w(\hat{x}) = P(e \mid \hat{x}).$ The reweighed samples are used to empirically approximate the expectation as 
$$\mathbb{E}_{x \sim P(X\mid e)}[P(Y \mid X=x)] \approx \frac{1}{|H|} \sum\nolimits_{\hat{x} \in H} P(Y \mid \hat{x}).$$

This model is guaranteed to infer the correct distribution, assuming 
\begin{enumerate}[label=(\arabic*)]
\item the emission model accurately encodes domain knowledge about noise, a property that is outside our models, 
\item the number of iterations is sufficiently large to ensure the Gibbs sampler converges (after discarding burn-in samples), which is a standard assumption for Gibbs sampling, and 
\item the full conditionals are accurately modeled by TabPFN,  which is reasonable as the original TabPFN framework is trained on a large number of simulated datasets and exhibits performance that equals or exceeds the best tabular learning methods.
\end{enumerate}

\section{Experimental Evaluation}
\begin{table*}[t]
\centering
\begin{tabular}{lcc}
& \multicolumn{2}{c}{Cleveland Heart Disease}  \\
\cmidrule(r){2-3} 
 & AUC-ROC $(\uparrow)$ & AUC-PR $(\uparrow)$\\
\midrule
TabPFN & \applygradient{0.85}{}{none}{.85}{0.87} 
        & \applygradient{0.83}{}{none}{0.83}{0.84} \\

TabRIM & \applygradient{0.87}{\textcolor{teal}{(+0.02)}}{none}{0.85}{0.87} 
        & \applygradient{0.84}{\textcolor{teal}{(+0.01)}}{none}{0.83}{0.84} \\
\end{tabular}
\begin{tabular}{cc}
 \multicolumn{2}{c}{Ljubljana Breast Cancer} \\
\cmidrule(r){1-2}
AUC-ROC $(\uparrow)$& AUC-PR $(\uparrow)$\\
\midrule

\applygradient{0.63}{}{none}{0.63}{0.74} 
& \applygradient{0.42}{}{none}{0.42}{0.54} \\

\applygradient{0.74}{\textcolor{teal}{(+0.11)}}{none}{0.63}{0.74} 
& \applygradient{0.54}{\textcolor{teal}{(+0.12)}}{none}{0.42}{0.54} \\
\end{tabular}
\caption{\textbf{RIMs with Gibbs sampling may outperform TabPFN} 
on tabular data with heavy observational noise. Scores are color-coded (the bluer, the better) and the differences are w.r.t.~TabPFN. (best viewed in color)}
\label{tab:results_TabRIM}
\end{table*}


\begin{table*}[t]
\centering
\setlength{\tabcolsep}{4pt}
\renewcommand{\arraystretch}{1.3}

\begin{tabular}{lcc}
    & \multicolumn{2}{c}{\textbf{ARC-AGI 1}} \\
    \cmidrule(r){2-3}
     & pass@1 $(\uparrow)$ & pass@2 $(\uparrow)$ \\
    \midrule
    SimRIM (aka TRM)   & \applygradient{40.5}{}{none}{40}{43.5}   & \applygradient{44.38}{}{none}{44.3}{47.5} \\
    RIMformer & \applygradient{43.25}{\textcolor{teal}{(+2.75)}}{none}{40}{43.5}  & \applygradient{47.13}{\textcolor{teal}{(+2.75)}}{none}{44.3}{47.5} \\
    RIMA      & \applygradient{42.5}{\textcolor{teal}{(+2.00)}}{none}{40}{43.5}  & \applygradient{47.5}{\textcolor{teal}{(+3.12)}}{none}{44.3}{47.5}  \\
 \end{tabular}
\begin{tabular}{cc}
 \multicolumn{2}{c}{\textbf{ARC-AGI 2}} \\
 \cmidrule(r){1-2}
 pass@1 $(\uparrow)$ & pass@2 $(\uparrow)$ \\
 \midrule
\applygradient{4.6}{}{none}{4}{10.0}           & \applygradient{7.4}{}{none}{7}{12.0} \\
\applygradient{5.83}{\textcolor{teal}{(+1.23)}}{none}{4}{10.0}         & \applygradient{7.1}{\textcolor{purple}{(-0.30)}}{none}{7}{12.0} \\
\applygradient{9.9}{\textcolor{teal}{(+5.30)}}{none}{4}{10.0} & \applygradient{11.3}{\textcolor{teal}{(+3.9)}}{none}{7}{12.0}
\end{tabular}
\begin{tabular}{cc}
 \textbf{Sudoku-Extreme} & \textbf{Maze-Hard} \\
 \cmidrule(l){1-1} \cmidrule(l){2-2}
 Accuracy $(\uparrow)$ & Accuracy $(\uparrow)$ \\
 \midrule
\applygradient{87.16}{}{none}{80}{89.5}           & \applygradient{85.3}{}{none}{85}{88} \\
\applygradient{80.21}{\textcolor{purple}{(-6.95)}}{none}{80}{89.5}           & \applygradient{87.7}{\textcolor{teal}{(+2.4)}}{none}{85}{88} \\
 \applygradient{89.34}{\textcolor{teal}{(+2.18)}}{none}{80}{89.5}  &  \applygradient{87.0}{\textcolor{teal}{(+1.7)}}{none}{85}{88}
\end{tabular}

\caption{\textbf{RIMs outperform Tiny Recursive Models (SimRIM) on symbolic reasoning benchmarks.} The test set performance (accuracy, pass@1, and pass@2) on Symbolic reasoning benchmarks (ARC-AGI-1, ARC-AGI-2, Sudoku-Extreme, and Maze-Hard), comparing a RIM with a simple identity Reweighter (SimRIM), with ones having more expressive Reweighers (RIMFormer and RIMA). Clearly, the RIMs with expressive Reweighters outperform the one with a simple identity Reweighter, underscoring the importance of reweighting for effective reasoning.Scores are color-coded (the bluer, the better) and the differences are w.r.t.~SimRIM/TRM. (best viewed in color)}
\label{tab:arc_results_combined}
\end{table*}

We aim to empirically investigate the performance and generalization of different RIMs across a range of tasks. To this end,  we conduct several experiments to address the following questions:
\begin{description}
    \item[(Q1)] How important is reweighting for neural reasoning tasks?
    \item[(Q2)] Which approach is more effective: dynamic or static reweighting?
    \item[(Q3)] Does a greater lookback lead to more accurate results?
    \item[(Q4)] Does the RIM framework generalize to tabular domains and provide robustness against observational noise?
\end{description}

\textbf{Domains.} 
To this end, we focused on two domains to evaluate instantiations of the RIMs: neural reasoning and tabular data. Within the neural reasoning domain, we evaluated on four benchmarks: ARC-AGI-1, ARC-AGI-2, Sudoku Extreme, and Maze-Hard.

ARC-AGI-1 consists of geometric puzzles designed to be solvable by humans while challenging current machine learning models \citep{chollet2019measure}. Each task provides 2--3 demonstration examples along with a test input and requires modeling global dependencies over 30×30 grids. ARC-AGI-2 expands on ARC-AGI-1 with a larger and more diverse set of tasks while preserving the few-shot setup \citep{chollet2025arc}. Both benchmarks require long-horizon reasoning across the grid, motivating the use of an attention block in the model backbone.

Sudoku Extreme contains complex 9×9 puzzles \citep{dillion2025tdoku}. The training set consists of 1{,}000 examples, whereas the test set contains 423{,}000 samples. Prior work has shown that a pure MLP-based backbone achieves the best performance on this benchmark \citep{jolicoeur2025less}, which we adopt for our experiments. 

Maze-Hard includes 30×30 mazes  \citep{lehnert2024beyond}, with 1{,}000 training and 1{,}000 test examples. Because these tasks involve long-range dependencies across the entire grid, neural modules for this task include an attention block. 

Together, these benchmarks span a spectrum of neural reasoning challenges, from small, fixed-context tasks to problems with large grids and long-horizon dependencies.

To evaluate our framework for tabular reasoning under uncertainty, we used two medical datasets from the UCI Machine Learning Repository: {\em Cleveland Heart Disease} and {\em Ljubljana Breast Cancer.} For each test set, we simulate severe data corruption by replacing 25\% of all feature values with random values.

\textbf{Models.}  In our experiments for the symbolic benchmarks, we evaluated two RIMs: RIMA and RIMformer, comparing them with SimRIM. Depending on the backbone used for the Solver and Generator, our models have between 5.5M and 13.6M parameters.

For the tabular reasoning experiments, we compared TabPFN and TabRIM. In TabRIM, the input is treated as a noisy realization of a latent clean state, which is refined by iteratively re-sampling each variable's values conditioned on the remaining variables' values. Equivalently, this can be viewed as a Gibbs sampling-based denoiser, where each conditional is defined via forward passes through a common TabPFN model. We generated 10 samples, after discarding 5 burn-in samples, and estimated the final answer as their mean predictive probability.

\textbf{Metrics.} To evaluate the performance on ARC-AGI, we report pass@1 and pass@2, i.e., the fraction of tasks where the top-1 or top-2 outputs match the ground truth. For Sudoku Extreme and Maze Hard, we report standard exact-match accuracy. For tabular reasoning, we quantified predictive performance in terms of AUC-ROC and AUC-PR.

    \textbf{Answer (Q1), Reweighting:} Table~\ref{tab:arc_results_combined} summarizes the performance of SimRIM, RIMformer, and RIMA on the 4 symbolic reasoning benchmarks. We observe that the RIMs with non-trivial Reweighters (RIMFormer and RIMA) consistently outperform the identity-reweighted RIM (SimRIM). These results suggest that reweighting is an important component of neural reasoning, as it enables the model to identify and prioritize high-signal latent trajectories. 
    
    \textbf{Answer (Q2), Dynamic vs. Static Weighting:} Table~\ref{tab:combined_results} presents an ablation study comparing dynamic and static reweighting strategies on the Sudoku Extreme benchmark. We evaluate the efficacy of our learned reweighting mechanism against static baselines and hybrid variants where reweighting is restricted to either the Solver or Generator outputs. The results indicate that fully dynamic, neural-driven reweighting consistently yields superior performance, suggesting that joint adaptation of both latent trajectories is essential for solving high-complexity reasoning tasks.
    
    \textbf{Answer (Q3), Lookback size:} 
    Table~\ref{tab:arc_results_combined} compares RIMA, having a lookback of size 1, with RIMformer, having a lookback of size $\max(N,T)$. 
    We observe that increasing the lookback window size does not yield better performance in all cases. While it improves performance on Maze-Hard, it reduces performance on Sudoku-Extreme.
    We hypothesize that this performance gain likely stems from the Maze-Hard's inherent demand for backtracking, where historical context is critical for navigating dead ends. Conversely, for Sudoku-Extreme, the most recent refinement step typically captures the sufficient statistics required for the latent state.
    So, the underperformance of a larger lookback RIMformer model might be attributed to overfitting, as the high-capacity Transformer Reweighter may over-parameterize the relatively simple constraints of these smaller instances.
    
    \textbf{Answer (Q4), Tabular reasoning:} Table~\ref{tab:results_TabRIM} compares TabRIM with TabPFN on two noisy medical diagnosis benchmark datasets. In both cases, the TabRIM outperforms direct inference. By explicitly modeling the transition between noisy and clean states through stochastic sampling, the model iteratively reasons about the distribution over true feature values before making a prediction

\begin{table}[ht]
\centering\small
\begin{tabular}{lr} 
\toprule
\textbf{Method} & \textbf{Sudoku Extreme} \\ 
\midrule
SimRIM (aka TRM)                            & \applygradient{87.16}{}{none}{80}{89.5} \\ 
 
$\left( \alpha_L,\alpha_H \right) = \left(0.4, 0.4 \right)$                          & \applygradient{80.26}{\textcolor{purple}{(-6.90)}}{none}{80}{89.5}     \\

$ \left( \alpha_L,\alpha_H \right) = \left(\theta_1, \theta_2 \right) $             & \applygradient{84.27}{\textcolor{purple}{(-2.89)}}{none}{80}{89.5}     \\
RIMA ($\tilde z_L$  only)          & \applygradient{87.84}{\textcolor{teal}{(+0.68)}}{none}{80}{89.5}      \\
RIMA ($\tilde z_H$ only)          &  \applygradient{87.57}{\textcolor{teal}{(+0.41)}}{none}{80}{89.5}      \\
RIMA           & \applygradient{89.34}{\textcolor{teal}{(+2.18)}}{none}{80}{89.5} \\

\bottomrule
\end{tabular}
\caption{\textbf{Reweighting Ablation: Static vs. Dynamic.} Test set accuracies on the Sudoku Extreme benchmark for RIMA (with full dynamic reweighting), and its variants with partial dynamic reweighting ($\tilde z_L$ only and $\tilde z_H$ only), learnable scalars $\left(\theta_1, \theta_2 \right)$ and fixed values. Clearly, the full reweighting outperforms other configurations. Scores are color-coded (the bluer, the better) and the differences are w.r.t.~SimRIM/TRM. (best viewed in color) }\label{tab:combined_results}
\end{table}

\section{Conclusion}
We introduced Recursive Inference Machines (RIMs) as a unifying framework for neural reasoning architectures. By formalizing inference dynamics as an explicit, iterative process, RIMs provide a global architectural perspective that exposes previously implicit design choices. This clarity enables principled generalizations—such as our RIMformer and $k$-lookback variants—that extend the capabilities of existing models. Our empirical evaluation across diverse symbolic and tabular reasoning domains demonstrates that RIMs outperform neural and tabular reasoning architectures.

The choice of the Reweighter in the presented instantiations leaves substantial room for further exploration. One promising direction is to employ an Extended Long Short-Term Memory (xLSTM) \citep{beck2024xlstm} as the Reweighter, enabling the model to capture the influence of previous reasoning steps through an explicit long-term memory mechanism. 
Exploring alternative neural architectures for other components may further improve the performance.
For instance, replacing the TRM's neural backbone with a Universal Transformer has been shown to yield strong improvements on reasoning tasks  \citep{gao2025universal}. While we do not evaluate this architecture in our experiments, it can be naturally combined with the proposed Reweighter within the RIM framework.
Further, while RIM goes beyond linear, Chain-of-Thought reasoning by explicitly reasoning about previously generated solutions and states, it could be extended to a Tree-of-Thoughts–style setup \citep{yao2023tree}, where multiple reasoning trajectories are explored concurrently and subsequently evaluated or reweighted. Such branching can be interpreted as parallel inference chains whose latent states are selectively retained or aggregated, offering a principled way to trade off exploration and computation.

Together, these directions position RIM as a modular path forward for designing the next generation of efficient, interpretable reasoning engines that bridge the gap between raw pattern matching and high-level symbolic manipulation.
While this transparency is critical for human oversight and trust-building, the increased reasoning capability may call for a more rigorous focus on safety during deployment.

\section{Acknowledgement}
We gratefully acknowledge support from the BMFTR project XEI (grant number 16IS24079B), the Cluster of Ex-
cellence ``Reasonable AI'' funded by the German Research
Foundation (DFG) under Germany’s Excellence Strategy
(EXC-3057), the DYNAMIC center funded by the LOEWE program of the Hessian Ministry of Science and Arts (grant number LOEWE1/16/519/03/09.001(0009)/98), and the Army Research Office (award W911NF2010224).


\bibliography{uai2026-template}

@article{jolicoeur2025less,
  title={Less is more: Recursive reasoning with tiny networks},
  author={Jolicoeur-Martineau, Alexia},
  journal={arXiv preprint arXiv:2510.04871},
  year={2025}
}

@article{wang2025hierarchical,
  title={Hierarchical Reasoning Model},
  author={Wang, Guan and Li, Jin and Sun, Yuhao and Chen, Xing and Liu, Changling and Wu, Yue and Lu, Meng and Song, Sen and Yadkori, Yasin Abbasi},
  journal={arXiv preprint arXiv:2506.21734},
  year={2025}
}

@article{beck2024xlstm,
  title={xlstm: Extended long short-term memory},
  author={Beck, Maximilian and P{\"o}ppel, Korbinian and Spanring, Markus and Auer, Andreas and Prudnikova, Oleksandra and Kopp, Michael and Klambauer, G{\"u}nter and Brandstetter, Johannes and Hochreiter, Sepp},
  journal={Advances in Neural Information Processing Systems},
  volume={37},
  pages={107547--107603},
  year={2024}
}

@article{roye2025tiny,
  title={Tiny Recursive Models on ARC-AGI-1: Inductive Biases, Identity Conditioning, and Test-Time Compute},
  author={Roye-Azar, Antonio and Vargas-Naranjo, Santiago and Ghai, Dhruv and Balamurugan, Nithin and Amir, Rayan},
  journal={arXiv preprint arXiv:2512.11847},
  year={2025}
}

@book{krantz2002implicit,
  title={The implicit function theorem: history, theory, and applications},
  author={Krantz, Steven George and Parks, Harold R},
  year={2002},
  publisher={Springer Science \& Business Media}
}

@book{koller2009probabilistic,
  title={Probabilistic graphical models: principles and techniques},
  author={Koller, Daphne and Friedman, Nir},
  year={2009},
  publisher={MIT press}
}

@article{asadulaev2025deep,
  title={Deep Improvement Supervision},
  author={Asadulaev, Arip and Banerjee, Rayan and Karray, Fakhri and Takac, Martin},
  journal={arXiv preprint arXiv:2511.16886},
  year={2025}
}

@article{gao2025universal,
  title={Universal Reasoning Model},
  author={Gao, Zitian and Chen, Lynx and Xiao, Yihao and Xing, He and Tao, Ran and Luo, Haoming and Zhou, Joey and Dai, Bryan},
  journal={arXiv preprint arXiv:2512.14693},
  year={2025},
  url={https://arxiv.org/abs/2512.14693}
}

@inproceedings{ross2011learning,
  title={Learning message-passing inference machines for structured prediction},
  author={Ross, Stephane and Munoz, Daniel and Hebert, Martial and Bagnell, J Andrew},
  booktitle={CVPR 2011},
  pages={2737--2744},
  year={2011},
  organization={IEEE}
}

@inproceedings{shi2023satformer,
  title={Satformer: Transformer-based unsat core learning},
  author={Shi, Zhengyuan and Li, Min and Liu, Yi and Khan, Sadaf and Huang, Junhua and Zhen, Hui-Ling and Yuan, Mingxuan and Xu, Qiang},
  booktitle={2023 IEEE/ACM International Conference on Computer Aided Design (ICCAD)},
  pages={1--4},
  year={2023},
  organization={IEEE}
}

@article{bhuyan2024neuro,
  title={Neuro-symbolic artificial intelligence: a survey},
  author={Bhuyan, Bikram Pratim and Ramdane-Cherif, Amar and Tomar, Ravi and Singh, TP},
  journal={Neural Computing and Applications},
  volume={36},
  number={21},
  pages={12809--12844},
  year={2024},
  publisher={Springer}
}

@article{wei2022chain,
  title={Chain-of-thought prompting elicits reasoning in large language models},
  author={Wei, Jason and Wang, Xuezhi and Schuurmans, Dale and Bosma, Maarten and Xia, Fei and Chi, Ed and Le, Quoc V and Zhou, Denny and others},
  journal={Advances in neural information processing systems},
  volume={35},
  pages={24824--24837},
  year={2022}
}

@article{liu2025trust,
  title={Trust, But Verify: A Self-Verification Approach to Reinforcement Learning with Verifiable Rewards},
  author={Liu, Xiaoyuan and Liang, Tian and He, Zhiwei and Xu, Jiahao and Wang, Wenxuan and He, Pinjia and Tu, Zhaopeng and Mi, Haitao and Yu, Dong},
  journal={arXiv preprint arXiv:2505.13445},
  year={2025}
}

@article{chollet2019measure,
  title={On the measure of intelligence},
  author={Chollet, Fran{\c{c}}ois},
  journal={arXiv preprint arXiv:1911.01547},
  year={2019}
}

@article{chollet2025arc,
  title={Arc-agi-2: A new challenge for frontier ai reasoning systems},
  author={Chollet, Francois and Knoop, Mike and Kamradt, Gregory and Landers, Bryan and Pinkard, Henry},
  journal={arXiv preprint arXiv:2505.11831},
  year={2025}
}

@misc{dillion2025tdoku,
  author       = {Dillon, T.},
  title        = {{Tdoku: A Fast Sudoku Solver and Generator}},
  year         = {2025},
  howpublished = {\url{https://t-dillon.github.io/tdoku/}},
  note         = {Accessed: 2025-01-31}
}

@article{lehnert2024beyond,
  title={Beyond a*: Better planning with transformers via search dynamics bootstrapping},
  author={Lehnert, Lucas and Sukhbaatar, Sainbayar and Su, DiJia and Zheng, Qinqing and Mcvay, Paul and Rabbat, Michael and Tian, Yuandong},
  journal={arXiv preprint arXiv:2402.14083},
  year={2024}
}

@incollection{pearl2022reverend,
  title={Reverend Bayes on inference engines: A distributed hierarchical approach},
  author={Pearl, Judea},
  booktitle={Probabilistic and causal inference: the works of Judea Pearl},
  pages={129--138},
  year={2022}
}

@book{pearl2014probabilistic,
  title={Probabilistic reasoning in intelligent systems: networks of plausible inference},
  author={Pearl, Judea},
  year={2014},
  publisher={Elsevier}
}

@article{geman1984stochastic,
  title={Stochastic relaxation, Gibbs distributions, and the Bayesian restoration of images},
  author={Geman, Stuart and Geman, Donald},
  journal={IEEE Transactions on pattern analysis and machine intelligence},
  number={6},
  pages={721--741},
  year={1984},
  publisher={IEEE}
}

@article{yao2023tree,
  title={Tree of thoughts: Deliberate problem solving with large language models},
  author={Yao, Shunyu and Yu, Dian and Zhao, Jeffrey and Shafran, Izhak and Griffiths, Tom and Cao, Yuan and Narasimhan, Karthik},
  journal={Advances in neural information processing systems},
  volume={36},
  pages={11809--11822},
  year={2023}
}

@article{hollmann2025accurate,
  title={Accurate predictions on small data with a tabular foundation model},
  author={Hollmann, Noah and M{\"u}ller, Samuel and Purucker, Lennart and Krishnakumar, Arjun and K{\"o}rfer, Max and Hoo, Shi Bin and Schirrmeister, Robin Tibor and Hutter, Frank},
  journal={Nature},
  volume={637},
  number={8045},
  pages={319--326},
  year={2025},
  publisher={Nature Publishing Group UK London}
}

@article{schuurmans2024autoregressive,
  title={Autoregressive large language models are computationally universal},
  author={Schuurmans, Dale and Dai, Hanjun and Zanini, Francesco},
  journal={arXiv preprint arXiv:2410.03170},
  year={2024}
}

@inproceedings{wang2022self,
  title={Self-Consistency Improves Chain of Thought Reasoning in Language Models},
  author={Wang, Xuezhi and Wei, Jason and Schuurmans, Dale and Le, Quoc V and Chi, Ed H and Narang, Sharan and Chowdhery, Aakanksha and Zhou, Denny},
  booktitle={The Eleventh International Conference on Learning Representations}
}

@inproceedings{ahuja2018lru,
  title={Lattice recurrent unit: Improving convergence and statistical efficiency for sequence modeling},
  author={Ahuja, Chaitanya and Morency, Louis-Philippe},
  booktitle={Proceedings of the AAAI Conference on Artificial Intelligence},
  volume={32},
  number={1},
  year={2018}
}

@inproceedings{cho2014gru,
  title={On the properties of neural machine translation: Encoder--decoder approaches},
  author={Cho, Kyunghyun and Van Merri{\"e}nboer, Bart and Bahdanau, Dzmitry and Bengio, Yoshua},
  booktitle={Proceedings of SSST-8, eighth workshop on syntax, semantics and structure in statistical translation},
  pages={103--111},
  year={2014}
}

@inproceedings{gu2024mamba,
  title={Mamba: Linear-time sequence modeling with selective state spaces},
  author={Gu, Albert and Dao, Tri},
  booktitle={First conference on language modeling},
  year={2024}
}

@inproceedings{peng2023rwkv,
  title={Rwkv: Reinventing rnns for the transformer era},
  author={Peng, Bo and Alcaide, Eric and Anthony, Quentin and Albalak, Alon and Arcadinho, Samuel and Biderman, Stella and Cao, Huanqi and Cheng, Xin and Chung, Michael and Derczynski, Leon and others},
  booktitle={Findings of the association for computational linguistics: EMNLP 2023},
  pages={14048--14077},
  year={2023}
}

@book{karpatne2022knowledge,
  title={Knowledge guided machine learning: Accelerating discovery using scientific knowledge and data},
  author={Karpatne, Anuj and Kannan, Ramakrishnan and Kumar, Vipin},
  year={2022},
  publisher={CRC Press}
}

@inproceedings{karanam2025unified,
  title={A unified framework for human-allied learning of probabilistic circuits},
  author={Karanam, Athresh and Mathur, Saurabh and Sidheekh, Sahil and Natarajan, Sriraam},
  booktitle={Proceedings of the AAAI Conference on Artificial Intelligence},
  volume={39},
  number={17},
  pages={17779--17787},
  year={2025}
}

\newpage

\onecolumn

\title{Recursive Inference Machines: A Unified View of Neural Reasoning\\(Supplementary Material)}
\maketitle


\appendix
\section{Hardware}

Experiments on Sudoku Extreme were conducted using two A100 GPUs (80GB). The experiments on Maze-Hard for the RIMA model were also performed on two A100 (80GB) with batch size of $128$. All other experiments used either four A100 GPUs (80GB) or four H100 GPUs. For ARC-AGI-1 and Sudoku Extreme, we reproduced the original TRM results using the authors’ codebase and reported a batch size of $768$. For ARC-AGI-2 and Maze-Hard, we report the results from \citep{jolicoeur2025less} due to computational constraints.

\section{Other Results}
\begin{table}[ht]
\centering\small

\begin{tabular}{lr} 
\toprule
\textbf{Method} & \textbf{Sudoku Extreme} \\ 
\midrule
SimRIM (aka TRM)                            & \applygradient{87.16}{}{none}{74}{89.5} \\ 
  $\left( \alpha_L,\alpha_H \right) = \left(0.8, 0.8 \right)$                          & \applygradient{82.76}{\textcolor{purple}{(-4.40)}}{none}{74}{89.5} \\

 $\left( \alpha_L,\alpha_H \right) = \left(0.6, 0.6 \right)$                          & \applygradient{85.29}{\textcolor{purple}{(-1.87)}}{none}{74}{89.5} \\
$\left( \alpha_L,\alpha_H \right) = \left(0.4, 0.4 \right)$                          & \applygradient{80.26}{\textcolor{purple}{(-6.90)}}{none}{74}{89.5}     \\
 $\left( \alpha_L,\alpha_H \right) = \left(0.2, 0.2 \right)$                          & \applygradient{74.52}{\textcolor{purple}{(-12.64)}}{none}{74}{89.5} \\

$ \left( \alpha_L,\alpha_H \right) = \left(\theta_1, \theta_2 \right) $             & \applygradient{84.27}{\textcolor{purple}{(-1.87)}}{none}{74}{89.5}     \\
RIMformer              & \applygradient{80.21}{\textcolor{purple}{(-6.95)}}{none}{74}{89.5}    \\  
RIMA ($\tilde z_L$  only)          & \applygradient{87.84}{\textcolor{teal}{(+0.68)}}{none}{74}{89.5}      \\
RIMA ($\tilde z_H$ only)          &  \applygradient{87.57}{\textcolor{teal}{(+0.41)}}{none}{74}{89.5}      \\
RIMA           & \applygradient{89.34}{\textcolor{teal}{(+2.18)}}{none}{74}{89.5} \\

\bottomrule
\end{tabular}
\caption{\textbf{Full Ablation Study on Static vs Dynamic reweighting} }\label{tab:combined_results2}
\end{table}

\end{document}